\documentclass[runningheads]{llncs}
\usepackage[T1]{fontenc}
\usepackage{booktabs}
\usepackage{amsmath}

\usepackage{xcolor}
\usepackage{booktabs}
\usepackage{cite}
\usepackage{marvosym}
\usepackage{graphicx}
\usepackage{indentfirst}
\usepackage{wasysym}
\usepackage{graphicx,verbatim}

\begin{document}
\title{Adaptive H\&E-IHC  information fusion staining framework based on feature extractor}
%

\author{Yifan Jia\textsuperscript{*}\inst{2}\textsuperscript{,}  \inst{4} \and Xingda Yu\textsuperscript{*}\inst{4} \and  Zhengyang Ji\inst{2}\textsuperscript{,} \inst{4} \and Songning Lai\inst{1}\textsuperscript{,} \inst{2} \and Yutao Yue\inst{1} \textsuperscript{,}\inst{2} \textsuperscript{,}\inst{3}\textsuperscript{(\Letter)}}  
\authorrunning{Anonymized Author et al.}
\institute{HKUST(GZ) \\
    \email{yutaoyue@hkust-gz.edu.cn} \and
    Deep Interdisciplinary Intelligence Lab \and
    Institute of Deep Perception Technology, JITRI \and 
    Shandong University \\ 
     }
\footnotetext[1]{\textsuperscript{*} Equal Contribution} 
\footnotetext[2]{Under Review}
   
\maketitle              
\begin{abstract}
Immunohistochemistry (IHC) staining plays a significant role in the evaluation of diseases such as breast cancer. The H\&E-to-IHC transformation based on generative models provides a simple and cost-effective method for obtaining IHC images. Although previous models can perform digital coloring well, they still suffer from (\romannumeral 1) coloring only through the pixel features that are not prominent in HE, which is easy to cause information loss in the coloring process; (\romannumeral 2) The lack of pixel-perfect H\&E-IHC groundtruth pairs poses a challenge to the classical L1 loss.To address the above challenges, we propose an adaptive information enhanced coloring framework based on feature extractors. We first propose the VMFE module to effectively extract the color information features using multi-scale feature extraction and wavelet transform convolution, while combining the shared decoder for feature fusion. The high-performance dual feature extractor of H\&E-IHC is trained by contrastive learning, which can effectively perform feature alignment of HE-IHC in high latitude space. At the same time, the trained feature encoder is used to enhance the features and adaptively adjust the loss in the HE section staining process to solve the problems related to unclear and asymmetric information. We have tested on different datasets and achieved excellent performance.Our code is available at \url{https://github.com/babyinsunshine/CEFF}


\end{abstract}
\section{Introduction}
Immunohistochemistry (IHC) staining is a widely used technique in pathology for visualizing common abnormal cells in tumors, which is crucial for developing precise treatment plans. However, traditional detection methods are both time-consuming and labor-intensive, with standard tissue pathology imaging involving in vivo tissue sampling, tissue fixation, tissue processing, section staining, microscopic observation, image capture, and image analysis~\cite{ref_1}. These factors hinder the widespread applicability of IHC staining in tissue pathology. With advancements in computer vision technology, researchers have applied computer vision techniques to the slide staining process (virtual staining), significantly improving detection efficiency and saving valuable time for patient treatment \cite{ref_2,ref_3,ref_4,ref_5}.

Existing virtual staining methods are mainly based on adversarial generation techniques. Liu et al. proposed PyramidPix2Pix~\cite{ref_6}, which applies Gaussian convolutions to image pairs and processes them at multiple scales, reducing the requirement for pixel-level precise alignment. Li et al. introduced a novel loss function designed to mitigate the negative impact of these inconsistencies on model performance~\cite{ref_6}. This loss function enables the model to better handle noise or low-quality data, thereby improving the robustness of the staining transformation. Li et al. also designed a multi-layer weak pathological consistency constraint, combined with an adaptive weight strategy and discriminator contrastive regularization loss, which significantly enhances the pathological consistency and realism of generated tissue slices~\cite{ref_7}.

Although the aforementioned studies have made significant advancements in the field of virtual staining, there are still several aspects that have not been fully addressed.  \romannumeral 1) Existing works mainly focus on j pixel information based stain generation task, overlooking the correspondence between potential staining grade labels of HE  and IHC slides, which is often a key factor that doctors consider during diagnosis. \romannumeral 2) The feature extraction methods used in current generator networks are limited and tend to overlook critical details, leading to poor detail in the generated IHC images.The information features in HE slides are not immediately apparent, which places a significant demand on the feature extraction capabilities of the generator network.  \romannumeral 3)The lack of pixel-perfect H\&E-IHC groundtruth pairs poses a challenge to the classical L1 loss.

To address the aforementioned issues, we make the following contributions: \textbf{1)} We propose the VMFE module, which employs multi-scale feature extraction and utilizes wavelet transform convolutions\cite{ref_8,ref_13,ref_14} for efficient extraction of staining information features, while incorporating a shared decoder for feature fusion. \textbf{2)} Inspired by contrastive learning\cite{ref_15,ref_16}, we pre-train feature encoders for HE (Hematoxylin and Eosin) and IHC (Immunohistochemistry) images, aiming to unsupervisedly align staining labels for HE and IHC images in the latent space. \textbf{3)} We leverage the trained feature encoders to enhance features and adaptively adjust the loss during the staining process for HE slides\cite{ref_17}, addressing issues related to unclear and asymmetric information. Finally, we conduct extensive testing across multiple datasets to validate the effectiveness of our method.
\section{Method}
\begin{figure}[t]
\includegraphics[width=\textwidth]{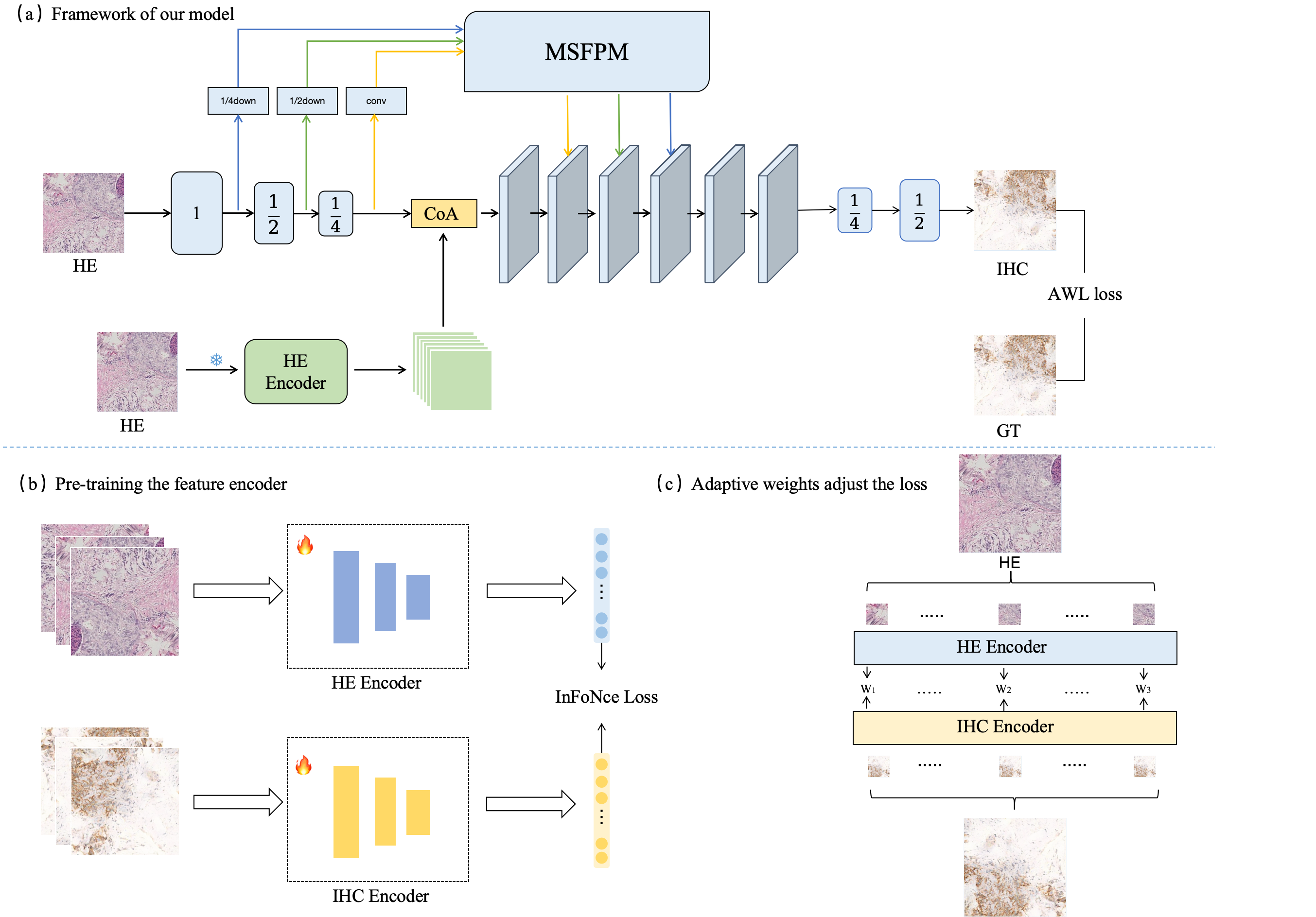}
\caption{Overview of the proposed framework.} \label{fig1}
\end{figure}

Figure~\ref{fig1} provides an overview of our proposed framework for adaptive IHC virtual staining. As shown in Figure \ref{fig1}(a), the architecture is centered on the Multi-Scale Modulated Feature Fusion Generator, which utilizes the \textbf{V}irtual \textbf{M}ulti-scale \textbf{F}eature \textbf{E}xtractor (VMFE) to process H\&E images. It achieves this by processing the downsampled features through the VMFE module and fusing them with later-layer feature maps to fully leverage information. Additionally, we use the Cross-Attention module (CoA) to fuse the feature maps obtained from the encoded H\&E images with those from the generator, providing more guidance for IHC image generation. Figure \ref{fig1}(b) highlights the pre-training process of the HE Encoder and IHC Encoder, where contrastive learning (using the InfoNCE loss function) trains the encoders to capture the semantic relationships between H\&E and IHC images. Figure \ref{fig1}(c) illustrates the adaptive L1 loss mechanism, which dynamically adjusts the loss weights based on the cosine similarity between the patch embedding vectors of the generated IHC image and the ground truth image, obtaining an adaptive weighted L1 (AWL) loss to address the non-strict symmetry issue between H\&E and IHC images, thereby improving staining accuracy.
\subsection{Multi-Scale Feature Extraction and Fusion}
Considering the issues of error propagation during sampling from low resolution to high resolution in a U-Net-like generator, as well as the insufficient information processing in large-scale skip connections, we propose a generator based on Virtual Multi-scale Feature Extraction (VMFE). The basic structure of this generator is resnet-6blocks, with VMFE replacing its downsampling component. VMFE primarily consists of wavelet convolution downsampling and a Multi-scale Sequential Feature Processing Module (MSFPM). For the input image \( X \), wavelet convolution-based downsampling layers produce multi-scale feature maps \( X_1 \), \( X_2 \), and \( X_3 \) with scales of 1, 1/2, and 1/4, respectively. These multi-scale feature maps have a larger receptive field.

Then, mimicking the coarse-to-fine approach of traditional U-Net, we sequentially input the multi-scale feature maps \( X_3 \), \( X_2 \), and \( X_1 \) into the MSFPM. The MSFPM utilizes a convolution-based Gated Recurrent Unit (GRU) to modulate the content between the previous activation \( h_{t-1} \) and the current input \( h_t \). This is because there exists an abstract temporal relationship among the feature maps obtained through downsampling. By using this module, we aim to enable the network to comprehensively consider the sequential relationship of each feature map. The hidden state update of the module can be simplified as:
\[
h_t = \text{MSFPM}(X_t, h_{t-1}),
\]
where \( X_t \) (\( t = 3, 2, 1 \)) denotes the input feature map, and \( h_t \) represents the output hidden state. Since \( X_3 \), \( X_2 \), and \( X_1 \) have different scales, we downsample each feature map to a 1/4 scale, denoted as \( \tilde{X}_t = \text{Downsample}(X_t, 1/4) \). Then, the module’s outputs \( h_3 \), \( h_2 \), and \( h_1 \) are respectively fused with the second, third, and fourth feature maps in the generator block via addition, i.e.,
\[
F'_k = F_k + h_{4-k}, \quad k = 2, 3, 4, \tag{5}
\]
where \( k \) denotes the index of the feature maps in the generator block (corresponding to \( k = 2, 3, 4 \) for the second, third, and fourth feature maps, respectively), \( F_k \) represents the original feature map, and \( F'_k \) denotes the fused feature map, thereby enhancing the model’s performance.
\subsection{Contrastive Learning Strategy of Dual Encoders}
In medical image processing, there must be an inherent connection between the images before and after staining, that is, they contain a large amount of the same semantic information. Based on this, we propose to use the method of contrastive learning to train two encoders, which respectively encode the images before and after staining. Pathological images contain a vast amount of complex information, and it is difficult to comprehensively capture all features using a single encoder. Therefore, we use two independent encoders for separate training to ensure that various features in the images can be fully mined, and to improve the comprehensiveness and accuracy of feature extraction.
 To measure the similarity and dissimilarity between encoded features, guiding the encoder to learn more discriminative and representative image features,We use the InfoNCE loss\cite{ref_18} function:
\begin{equation}
L_{NCE} = -\frac{1}{N}\sum_{i = 1}^{N}\log\frac{\exp(s(z_i, z_{i}^{+})/\tau)}{\sum_{j = 1}^{M}\exp(s(z_i, z_{j})/\tau)},
\end{equation}
where \(z_i\) and \(z_i^+\) represent the feature vector of the \(i\)-th sample and its positive counterpart, \(s(z_i, z_j)\) is the cosine similarity score, and \(\tau\) is the temperature parameter. \(N\) is the batch si and \(M\) is the number of negative samples.
To prevent overfitting during training, we introduce an L2 regularization term:

\begin{equation}
L2 = \lambda\sum_{w\in\theta}|w|_{2}^{2},
\end{equation}
where \(\lambda\) is the regularization strength and \(\theta\) represents the set of model parameters.

Finally, our total loss formulations are as follows:
\begin{equation}
L = L_{NCE} + \lambda \sum_{w\in\theta} |w|_{2}^{2}.
\end{equation}

\subsection{Cross-Attention Feature Fusion between Encoder and Generator}
To leverage the information captured by the trained H\&E encoder---owing to the use of contrastive loss, which encodes mutual information between H\&E and IHC images---we propose a cross-attention fusion module. This module integrates a feature map from a specific layer of the encoder with the first feature map of the generator block to guide the staining process.

Given the generator feature map \( F_{\text{gen}} \in \mathbf{R}^{B \times C \times H \times W} \) and the encoder feature map \( F_{\text{enc}} \in \mathbf{R}^{B \times C \times H \times W} \), we generate queries \( Q \), keys \( K \), and values \( V \) via 1\(\times\)1 convolutions, followed by reshaping into \( \mathbf{R}^{B \times N \times d} \), where \( N = H \times W \) and \( d \) is the feature dimension. The fusion process is defined as follows:

The output feature map is computed as:
\begin{equation}
F_{\text{out}} = F_{\text{gen}} + \alpha \cdot \text{BN} \left( W_{\text{out}} * \left( \text{softmax} \left( \frac{Q K^\top}{\sqrt{d}} \right) V \right) \right),
\end{equation}
where \( W_{\text{out}} \) denotes the 1\(\times\)1 convolution weight, \( * \) represents the convolution operation, \(\text{BN}\) is batch normalization, and \( \alpha \) is a hyperparameter controlling the fusion strength. Through this approach, the generator effectively incorporates the encoder's information, improving the accuracy of generating IHC images from H\&E images.
\subsection{Adaptive L1 Loss}
Due to the non-strict symmetry between H\&E images and IHC images, we adapt the L1 loss weight by leveraging the encoding information from the IHC encoder. The generated image and the ground truth are divided into multiple patches, and the cosine similarity of the corresponding patches’ embedding vectors, after passing through the IHC encoder, is computed. The adaptive L1 loss is defined as:

\begin{equation}
L_{1} = \sum_{i=0}^{n-1} \left( \alpha + \beta \cdot \text{Sim}_i \right)/n
\end{equation}

where \( \text{Sim}_i \) is the cosine similarity between the embedding vectors of the corresponding patch pair, and lower similarity often indicates poor symmetry, thus reducing the L1 loss weight.

\begin{table}[t]
\centering 
\caption{Comparative Performance Evaluation on Histopathology Datasets}
\label{tab:enhanced_performance}
\small 
\begin{tabular}{@{}lcrc@{\hspace{2em}}lcrc@{}}
\toprule 
\multicolumn{4}{c}{\textbf{HER2Bci}} & \multicolumn{4}{c}{\textbf{ERMist}} \\
\cmidrule(r){1-4} \cmidrule(l){5-8}
Method & \multicolumn{3}{c}{Metrics} & Method & \multicolumn{3}{c}{Metrics} \\
\cmidrule(r){2-4} \cmidrule(l){6-8}
 & PSNR↑ & SSIM↑ & FID↓ &  & PSNR↑ & SSIM↑ & FID↓ \\
\midrule 
CycleGAN & 14.201 & 0.424 & 63.7 & CycleGAN & 11.900 & 0.181 & 88.7 \\
CUT & 17.322 & 0.438 & 65.0 & CUT & 12.030 & 0.183 & 47.1 \\
PyramidP2P &\textcolor{blue}{21.160} & 0.477 & 80.1 & PyramidP2P & 12.100 & 0.191 & 80.8 \\
ASP & 17.869 & 0.492 & \textcolor{blue}{54.3} & ASP & 13.890 & 0.206 & 41.2 \\
ESI & 19.132 & \textcolor{blue}{0.499} & \textcolor{red}{50.1} & ESI & \textcolor{blue}{13.900} & \textcolor{blue}{0.209} & \textcolor{blue}{34.9} \\
Ours &\textcolor{red}{ 21.380} & \textcolor{red}{0.504} & 94.1 & Ours & \textcolor{red}{15.562} & \textcolor{red}{0.243} & \textcolor{red}{30.9} \\
\addlinespace 
\toprule 
\multicolumn{4}{c}{\textbf{PRMist}} & \multicolumn{4}{c}{\textbf{Ki67Mist}} \\
\cmidrule(r){1-4} \cmidrule(l){5-8}
Method & PSNR↑ & SSIM↑ & FID↓ & Method & PSNR↑ & SSIM↑ & FID↓ \\
\midrule 
CycleGAN & 12.990 & 0.187 & 78.6 & CycleGAN & 12.917 & 0.201 & 100.8 \\
CUT & 13.560 & 0.192 & 53.2 & CUT & 13.697 & 0.212 & 53.1 \\
PyramidP2P & 14.430 & 0.224 & 79.2 & PyramidP2P & 13.987 & 0.248 & 89.8 \\
ASP & 14.330 & 0.216 & \textcolor{blue}{44.5} & ASP & 14.824 & 0.241 & \textcolor{blue}{50.9} \\
ESI & \textcolor{blue}{15.936} & \textcolor{blue}{0.248} & \textcolor{red}{34.2} & ESI & \textcolor{blue}{16.093} & \textcolor{blue}{0.262} & \textcolor{red}{31.1} \\
Ours & \textcolor{red}{15.990} & \textcolor{red}{0.290} & 93.7& Ours & \textcolor{red}{16.210} & \textcolor{red}{0.316} & 107.6 \\
\bottomrule 
\end{tabular}
 
\vspace{0.5em}
\footnotesize 
The red value indicates the best performance case. Blue indicates the second-best performance case.
\end{table}
\section{Experiments}
\subsection{Experimental Setup}
\subsubsection{Datasets}
In this study, we selected two key datasets: the Breast Cancer Immunohistochemistry (BCI)~\cite{ref_6} Challenge dataset and the MIST dataset~\cite{ref_9}. The BCI dataset comprehensively covers different levels of HER2 expression, providing a rich data foundation for in - depth research on the characteristics related to HER2 expression. The MIST dataset, on the other hand, contains immunohistochemical staining data for HER2, PR, ER, and Ki67, presenting information on breast cancer - related indicators from multiple dimensions. Our division of the test set and training set is consistent with that in the original paper.
\subsubsection{Experimental Details}
Our model was trained on an NVIDIA RTX 3090 GPU. For both the encoder and the model, we employed the Adam optimizer. The encoder was trained for 300 epochs with a batch size of 64, while the model was trained for 100 epochs with a batch size of 1. We randomly cropped the images to a size of 512×512 for training.The fusion strength was set to 0.2, while the parameters $\alpha$ and $\beta$ of the adaptive L1 loss were both set to 50.

\subsubsection{Evaluation Methods}
To comprehensively evaluate the model, we adopted multiple metrics. PSNR measures the distortion between generated and real images, with a higher value indicating better quality. SSIM assesses structural similarity, closer to 1 meaning more similar structures and better aligning with human vision. FID quantifies the difference between the distributions of generated and real images, with a lower value denoting better quality and diversity. 
\begin{figure}[t]
\includegraphics[width=\textwidth]{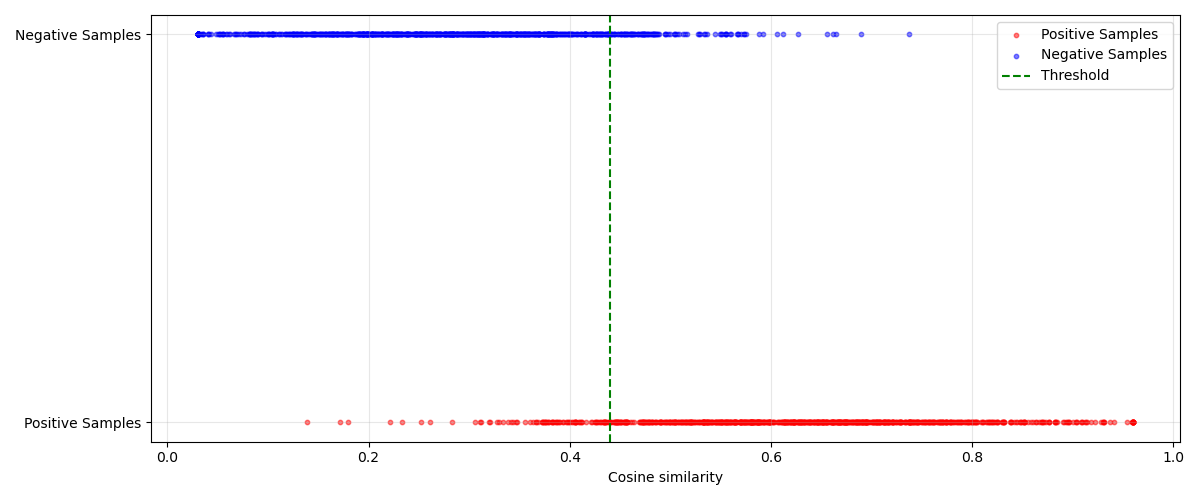}
\caption{Encoder performance analysis on BCI dataset.}
\label{fig2}
\end{figure}

\begin{table}[b]
\centering
\caption{Ablation Study on ERMist Dataset}
\label{tab:ablation}
\begin{tabular}{|l|c|c|c|c|}
\hline
Configuration                            & PSNR $\uparrow$ & SSIM $\uparrow$ & FID $\downarrow$  \\
\hline

Without Multi-Scale Feature Extraction    & 15.357            & 0.235            & 66.66                         \\
Without Cross-Attention Feature Fusion    & 15.423            & 0.237            & 67.73                  \\
Without Adaptive L1 Loss & 15.437            & 0.240            & 89.77                 \\
Full Model (All Methods)                 & 15.562            & 0.243            & 30.9                \\
\hline
\end{tabular}
\end{table}
\begin{figure}[t]
\includegraphics[width=\textwidth]{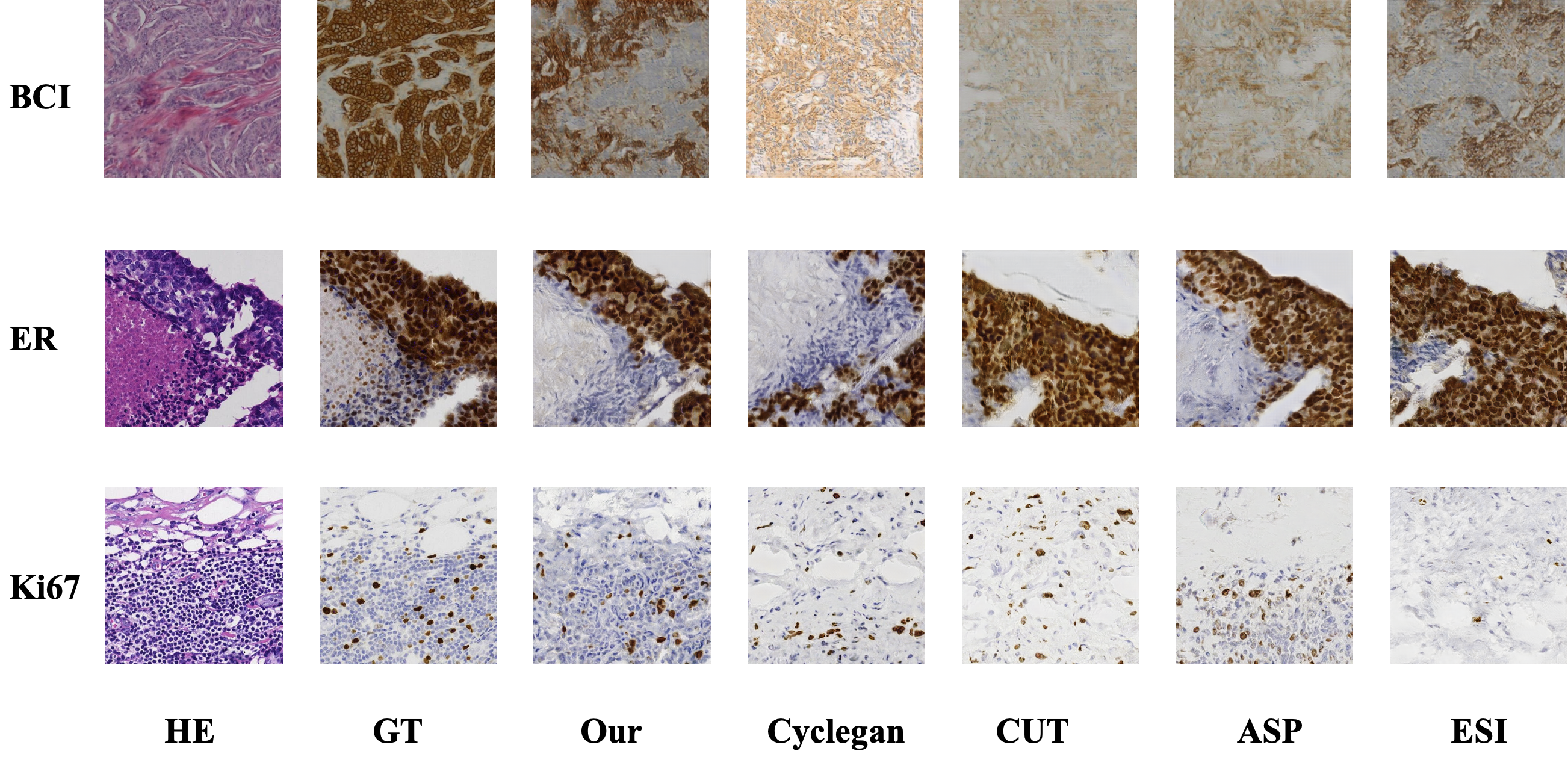}
\caption{Visualize different methods on different dataset images.} \label{fig3}
\end{figure}
\subsection{Comparative Experiments}
\subsubsection{The performance of the dual encoder.}
Dual Encoders aim to capture the consistency of paired H\&E and IHC images using contrastive learning. In this section, we show the effectiveness of the dual encoder. We test the performance of the dual encoder by constructing paired pairs of positive samples and unpaired pairs of negative samples. As shown in Figure \ref{fig2}, after coding and calculating the similarity between H\&E and IHC, we can find that there is a clear boundary between the similarity of positive sample pairs and negative sample pairs. When the similarity boundary is 0.44, the recognition accuracy of positive and negative sample pairs reaches up to 92.37\%. When the similarity is less than 0.44, HE and IHC are mostly unpaired data, and when the similarity is greater than 0.44, it is mostly paired data.
\subsubsection{Comparison with State-of-the-arts.}
Table~\ref{tab:enhanced_performance} summarizes the quantitative comparison results on the BCI dataset. We compared our proposed method with the following five methods: CycleGAN~\cite{ref_10}, Cut~\cite{ref_11}, Pyramid Pix2Pix~\cite{ref_6}, ASP~\cite{ref_12}, and ESI~\cite{ref_7}.Our proposed method achieved competitive performance across various datasets, attributable to the integration of contrastive learning, multi-scale feature fusion, and adaptive L1 loss. On the MIST dataset, which includes multiple IHC markers (HER2, PR, ER, Ki67), our method maintained its superiority, particularly in the PSNR and SSIM metric.Our method performs slightly worse on the FID metric, but also achieves sota results on some datasets.Fig.~\ref{fig3} illustrates representative IHC images generated from H\&E inputs. Compared to the baselines, our method produced sharper edges, richer textures, and more accurate protein expression patterns, especially in regions with complex tissue morphology.
\subsubsection{Ablation Experiments}
To evaluate the contribution of each component in our proposed framework, we conducted an ablation study on the BCI Challenge dataset, as shown in Table \ref{tab:ablation}. We used the full model, incorporating all components, as the performance reference. Replacing the VMFE module with the original network led to a decline in the ability to preserve pathological details, resulting in reduced overall performance. Removing the cross-attention feature fusion module decreased the utilization efficiency of information between the encoder and generator, affecting staining accuracy. Excluding the adaptive L1 loss exacerbated the issue of image asymmetry, further degrading performance. These results underscore the importance of each component, demonstrating that their combined effect is crucial for achieving superior H\&E-to-IHC virtual staining performance.
\section{Conclusion}
We propose an adaptive IHC virtual staining method framework using contrastive-encoding feature fusion. By aligning H\&E and IHC features via dual-branch contrastive learning, enhancing structural consistency with cross-attention fusion, and mitigating asymmetry with a dynamic L1 loss, our method outperforms existing approaches. Experiments and ablation studies validate its effectiveness in improving staining quality and detail preservation. This framework offers a promising tool for rapid, cost-effective pathological diagnosis with potential clinical impact.

\subsubsection{Acknowledgments.}
This work was supported by Guangzhou-HKUST(GZ) Joint Funding Program(Grant No.2023A03J0008), Education Bureau of Guangzhou Municipality.

\end{document}